\documentclass[10pt,twocolumn,letterpaper]{article}

\usepackage{iccv}
\usepackage{times}
\usepackage{epsfig}
\usepackage{graphicx}
\usepackage{amsmath}
\usepackage{amssymb}
\usepackage{booktabs}

\usepackage{amsmath,amsfonts,bm}

\def\eqref#1{equation~\ref{#1}}

\def\1{\bm{1}}

\def\rvx{{\mathbf{x}}}
\def\rvy{{\mathbf{y}}}
\def\rvz{{\mathbf{z}}}

\DeclareMathAlphabet{\mathsfit}{\encodingdefault}{\sfdefault}{m}{sl}
\SetMathAlphabet{\mathsfit}{bold}{\encodingdefault}{\sfdefault}{bx}{n}

\def\sR{{\mathbb{R}}}

\def\sX{{\mathbb{X}}}
\def\sY{{\mathbb{Y}}}
\def\sZ{{\mathbb{Z}}}

\newcommand{\E}{\mathbb{E}}

\newcommand{\KL}{D_{\mathrm{KL}}}

\DeclareMathOperator*{\argmax}{arg\,max}

\usepackage[breaklinks=true,bookmarks=false]{hyperref}

\iccvfinalcopy 

\def\iccvPaperID{3560} 

\ificcvfinal\fi
\begin{document}

\title{Mode matching in GANs through latent space learning and inversion}

\author{Deepak Mishra$^\ast$, Prathosh AP\thanks{Equal contribution.} , Aravind Jayendran, Varun Srivastava, and Santanu Chaudhury \\
Indian Institute of Technology Delhi\\
New Delhi, India\\
{\tt\small deemishra21@gmail.com}
}

\maketitle

\begin{abstract}
Generative adversarial networks (GANs) have shown remarkable success in generation of unstructured data, such as, natural images. However, discovery and separation of modes in the generated space, essential for several tasks beyond naive data generation, is still a challenge. In this paper, we address the problem of imposing desired modal properties on the generated space using a latent distribution, engineered in accordance with the modal properties of the true data distribution. This is achieved by training a latent space inversion network in tandem with the generative network using a divergence loss. The latent space is made to follow a continuous multimodal distribution generated by reparameterization of a pair of continuous and discrete random variables. In addition, the modal priors of the latent distribution are learned to match with the true data distribution using minimal-supervision with negligible increment in number of learnable parameters. We validate our method on multiple tasks such as mode separation, conditional generation, and attribute discovery on multiple  real world image datasets and demonstrate its efficacy over other state-of-the-art methods.

\end{abstract}

\section{Introduction}
Separation of data modes in a generative model amounts to dividing the data manifold into regions representing classes or distinguishable attributes and sampling new points from them \cite{jaakkola1999exploiting,hastie1996discriminant}. Several learning tasks such as conditional generation, data augmentation, supervised and semi-supervised classification, class balancing and clustering~\cite{hastie2009unsupervised,Hinton06,hinton2006reducing,bengio2013representation,radford2015unsupervised,kingma2014semi,cheung2014discovering} can be achieved with generative models that separate the data modes.  Among modern generative models, Generative Adversarial Network (GAN)~\cite{goodfellow2014generative} is a highly successful generative neural network that learns mappings from arbitrary latent distributions to complex real-world distributions \cite{radford2015unsupervised,isola2017image,van2016wavenet,odena2016conditional,denton2015deep}.

However, it has been observed in practice that GANs are unable to reproduce all the modes in the true data in their raw formulation, a phenomenon infamous as the problem of \textit{mode collapse} in GANs.
\cite{wu2017gp,karras2017progressive,brock2018large}. 
This effectively means that the generator ends up fooling the discriminator not by discovering all the data modes but by traversing within a limited support of the data distribution \cite{salimans2016improved}. The severity of the degeneration in the produced modes is more prominent in case when the data modes are skewed, which is a common phenomenon in many domains such as medical-scans, genomes and surveillance.  
Further, many applications demand that the data is conditionally generated from a particular data mode. For instance, when generative models are used for data augmentation to solve a classification problem with severe class-imbalance, one would expect to sample from the data-mode corresponding to the sparser class.  Thus it is desirable that in a GAN all the modes of the generated data are well-separated and matched with the true data modes. In this work, we propose to address these issues by latent space engineering.

\subsection{Prior Art}
\label{related}

Considerable amount of literature exists on addressing the problem of mode discovery within the GAN framework. These works can be broadly categorized under the following two buckets -  the ones that alter GAN's structure and/or the learning objective for better mode discovery \cite{arjovsky2017wasserstein,metz2016unrolled,mao2017least,salimans2016improved,ghosh2018multi,yu2018mixture,deng2017structured} and those which explicitly encourage GANs to discover diverse data modes through latent space inversion  \cite{srivastava2017veegan, dumoulin2016adversarially,donahue2016adversarial,chen2016infogan,springenberg2015unsupervised,mukherjee2018clustergan}. We will briefly review such methods in this section and enlist our contributions subsequently. 

\subsubsection{Modified GAN learning for mode discovery}

Salimans et. al. \cite{salimans2016improved} proposed a series of tricks namely feature matching, minibatch discrimination and historic averaging to stabilize the GAN training and avoid mode collapse. In WGAN~\cite{arjovsky2017wasserstein} and LSGAN~\cite{mao2017least} earth-mover and Pearson divergence distances are respectively used in the GAN objective instead of the JS-divergence, to induce stability in training which results in better mode discovery.  In~\cite{metz2016unrolled}, the generator objective is optimized by including the unrolled states of the discriminator, to address mode dropping. Drawing motivations from multi-agent learning, MADGAN \cite{ghosh2017multi} and GANMM \cite{yu2018mixture} employ the idea of using multiple generators and show that overall scheme performs similar to a mixture model with each generator learning one data mode.



\subsubsection{Latent posterior estimators}
In these class of methods, the basic theme is to learn mappings simultaneously from the latent to the data space and backwards, auto-encoding the latent/noise space. This idea is explored in multiple contexts including regularization of GAN, representation learning and mode discovery by methods such as  InfoGAN \cite{chen2016infogan},  Adversarially learned inference (ALI) \cite{dumoulin2016adversarially}, Bidirectional GAN (BiGAN) \cite{donahue2016adversarial}, VEEGAN \cite{srivastava2017veegan}, structured GAN \cite{deng2017structured}, CaTGAN \cite{springenberg2015unsupervised} and ClusterGAN \cite{mukherjee2018clustergan}. Motivated by the idea of regularized information maximization \cite{krause2010discriminative}, InfoGAN and CaTGAN train the GAN along with a regularization term to maximize the mutual information between the generated data and parts of the factorized latent space. The objective of VEEGAN \cite{srivastava2017veegan} is to map both the true and the generated data to a fixed Normal distribution in a variational framework to reduce mode collapse. In ALI and BiGAN, an encoder is learned that would map the data space to the latent space. ClusterGAN, employs a mixture of continuous and discrete latent space with inversion and clustering-loss to achieve clustering in the latent space post training.

\subsection{Problem setting}
While all the aforementioned methods offer several advantages in their own respects including stabilization of GAN training, reduction of mode collapse and  disentangled latent representation, there remains a major issue which is unaddressed -  All the aforementioned methods employ a fixed latent distribution, either a unimodal Normal (e.g., WGAN, ALI etc.) or a factored multimodal distribution (e.g., InfoGAN, ClusterGAN etc.) with uniform mode priors. Both of these distributional choices for the latent space result in a mismatch between the modes of generated and true data distributions, when the true data distribution has a non-uniform mode prior \cite{kim2018disentangling,ghosh2017multi}. Such a scenario occurs typically in the cases where every data mode represents a class and there is sever imbalance in class distribution. Thus we intend to formulate and address the following question in this work - \textbf{\textit{How to choose/learn the latent space distribution in a GAN framework such that modal properties of the true and the generated space are matched?}}

\subsection{Motivation and contributions}

In a latent variable generative model, since the data-likelihood is obtained by marginalizing the product of the conditional likelihood and the latent prior, it is natural to have a multimodal prior on the latent variable, to make the data distribution multimodal. This is the case with conventional generative models such as Gaussian Mixture Models, where the latent variable follows a multinoulli distribution, chosen to denote the expected number of modes in the data. Motivated by this, we first propose to employ a learnable multimodal latent space in a GAN framework. We show that, while having a multimodal latent space is necessary to generate multimodal data, it is not sufficient since the generative model can still collapse data modes ignoring the latent variable, especially when the data modes are severely skewed and/or not `well separated'. Thus we additionally propose to employ a data-to-latent inversion strategy and enforce the model to infer the modal posterior of the latent space, back from the generated data. Finally we show that the modal properties of the true data distribution are imposed on the generated distribution by learning the latent mode priors through minimal-supervision of the latent space inverter. The core contributions of our work are enlisted below:




\begin{enumerate}
	
	\item {Construction, sampling and learning of a multi-modal latent distribution by  reparameterization of a mixture of continuous and discrete random variables that could be employed in any GAN framework.}
	

	\item {Imposition of modal properties of the latent space on the generated space using two-stage latent inversion.} 
	
	
	
	\item {Leveraging the latent inverter with a divergence loss between the true and the estimated priors, for learning the mode priors of the latent distribution to follow true-data distribution using minimal-supervision.} 
	
	
	\item {Mathematical formulation of the problem of mode matching by treating neural networks as function approximators and deriving exact loss terms that are devoid of approximate/variational inferences and bounds.}
	
	\item {Validation of the proposed method on three-real world datasets with standard experiments along with proposal of new experiments with skewed data modes.}
	
\end{enumerate}





\section{Proposed Method}

\begin{figure}[!t]
\centering
{\includegraphics[width=\linewidth]{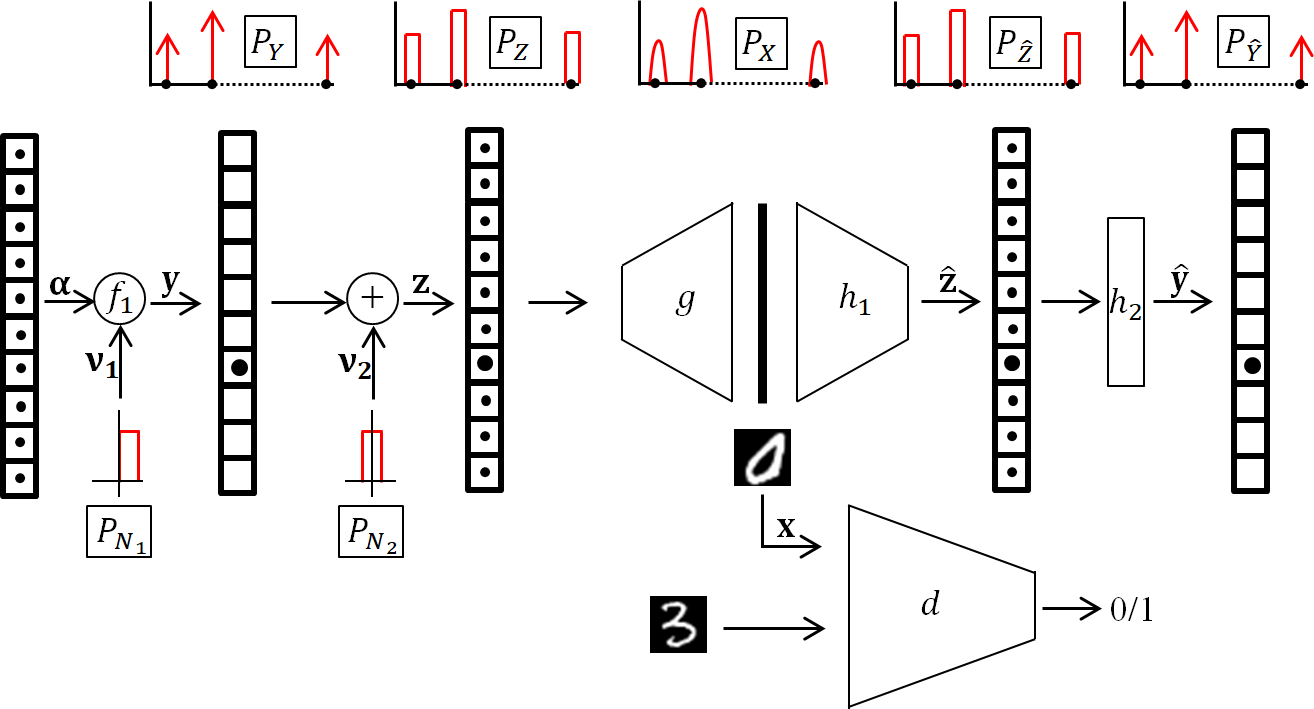}}
\caption{Illustration of the proposed method. Latent vector $\rvz$ is sampled from an additive mixture of a continuous $P_{N_2}(\boldsymbol{\nu_2})$ and discrete $P_Y(\rvy)$ distribution, resulting in a multimodal distribution $P_{Z}$. Generator $g(\rvz)$ tries to mimic the true data distribution $P_{X}$ with the help of discriminator $d$. The inversion network $h_2(h_1(g(\rvz)))$ inverts the generation process to ensure the matching of modal properties of generating and latent distributions. Mode priors of the latent space is encoded in $\rvy$ by reparametrizing a known distribution $P_{N_1}({\nu_1})$ using a learnable vector $\boldsymbol{\alpha}$. The priors are learned for an accurate matching of the latent and  data generating distributions with that of the true data distribution.}
\label{bd}
\end{figure}

We describe our method in four parts - In the first part, we present a method to parameterize and sample from a multimodal continuous distribution that forms the latent space. In the second part, we formulate the procedure to match the modes of the latent and the data generating distribution using latent space inversion. In the third part, we realize mode matching in a GAN framework. In the last part, we detail the procedure to learn the mode priors of the latent space so that modal properties of the latent and true data distributions are matched. 

\subsection{Noise engineering - Latent space construction}

Engineering the latent space in accordance with the true data distribution is the first problem of interest. In this work, we target matching two modal properties namely  1) the number of modes and 2) modal mass or mode priors. We desire the latent distribution to have same number of modes with their corresponding priors as the true data distribution. Let the latent space be represented by $\sZ$ and $P_{Z}$ denote its distribution. We define $P_{Z}$ as a multimodal distribution with $M$ modes, if its support, ${\sZ}$ is a disconnected set, which is a union of $M$ non-empty pairwise disjoint connected open subsets $(\sZ_i, i\in\{0,1,...,M-1\})$, each representing a mode.  
One can construct such multimodal distribution by taking an additive mixture of a generalized discrete or multinoulli distribution and a compact-support continuous distribution such as uniform distribution. Let $\rvy \sim P_{Y}$  and $\boldsymbol{\nu_2} \sim P_{N_2}$ denote samples drawn from such discrete  and continuous distributions, respectively. Then the latent space $\rvz$ is given as 
\begin{equation}
    \rvz =  \rvy + \boldsymbol{\nu_2}
\end{equation}
As shown in Figure~\ref{bd}, this results in a multimodal continuous distributions with disconnected modes as $P_Z = P_Y * P_{N_2}$, where $*$ denotes the convolution product. Note that the support of $P_{N_2}$ is chosen in such a way that the modes of $P_Z$ are disjoint.  In $P_Z$, the number and the mass of the modes are obtained from discrete component $(P_Y)$ and the continuous component $(P_{N_2})$ ensures the variability.  The discrete component $\rvy\sim P_Y$ can also be interpreted as an indicator of the modes of $\rvz$. Formally,  $\rvy:= i \quad\forall {z\in\sZ_i}$,  which implies 
\begin{equation}
    \int_{ \sZ_i }P_{Z} ~d\rvz = P_{Y}(\rvy = i )
    \label{mode}
\end{equation}

Thus, a natural choice for  $P_Y$ would be to follow the true data distribution such that it has same number of modes and priors as the true data distribution. This would require the knowledge of the modal properties of the true data, however, in real-world scenarios, often the number of modes are known (typically the number of distinct classes in the data) but not the mode priors (class imbalance). For instance, in a typical medical-scan classification problem, one would know that there are K-classes (E.g., benign and malignant) but not the class-prior. It is, therefore, necessary to construct $P_Y$ with the knowledge of number of modes, with the mode priors being learned during training. This is not possible in conventional settings as in InfoGAN, ClusterGAN~\cite{chen2016infogan,mukherjee2018clustergan}, since the learning of mode priors would require backpropogation to flow through the stochastic sampling ($\rvy \sim P_Y$) node. We, therefore, propose to reparametrize a second continuous uniform distribution, $P_{N_1}$, using a vector $\boldsymbol{\alpha}$ to construct the desired $P_Y$, as explained below.

Let $\boldsymbol{\alpha} = [\alpha_0,\alpha_1,.....,\alpha_{M-1}]^T$, $\alpha_i \in \mathbb{R}$ be an arbitrary vector and   $\nu_1 \sim P_{N_1}(\nu_1) = \mathbb{U}(0,1) $. We define a function, $f\left(\boldsymbol{\alpha}, \nu_1\right): \sR^M\times \sR \rightarrow \sR^M$ reparameterizing  $P_Y$ as follows.
\begin{equation}
    f_i\left(\alpha_i, \nu_1\right) = \begin{cases}\sigma_h\left( a_i-\nu_1\right) - \sigma_h\left(a_{i-1}-\nu_1\right); &i\neq 0\\
    \sigma_h\left( a_i-\nu_1\right);  &i=0
    \end{cases}
    \label{f1}
\end{equation}
where$f_i$ is the $i^{th}$ element of $f$, $\sigma_h$ is a unit step  function and $a_i$ is given as
\begin{equation}
    a_i = \frac{1}{{\sum_{k}{e^{\alpha_k}}}}\sum_{j = 0}^{i}{e^{\alpha_i}}
\end{equation}

With these definitions we show that one can reparametrize a uniform distribution using $\boldsymbol{\alpha}$ and $f$ to obtain a multinoulli distribution. 

\textbf{Lemma 1.} Define $\rvy =: \argmax_{i\in \{0,..,M-1\}} f_{i}$, then $\rvy$ follows a multinoulli distribution $P_Y$  with
\begin{equation}
    P_Y(\rvy=i) = \frac{e^{\alpha_i}}{\sum_{k}{e^{\alpha_k}}}
    \nonumber
\end{equation}

\textbf{Proof:} Since $\sigma_h$ is a unit step function, $f$ is the first order difference or discrete Dirac delta function positioned at $a_i$.
Now by definition,
\begin{equation}
    P_Y(\rvy=i) = P(f_{i} \neq 0)
    \label{prior}
\end{equation}
From \eqref{f1}, we can see that $f_{i}$ becomes non-zero only for $a_{i-1}\leq\nu_1\leq a_i$, therefore,

\begin{align}
P_Y(\rvy=i) &= P_{N_1}(a_{i-1}\leq\nu_1\leq a_i)\\
     &= \int_{a_{i-1}}^{a_i}P_{N_1}(\nu)d\nu = a_i - a_{i-1} = \frac{e^{\alpha_i}}{\sum_{k}{e^{\alpha_k}}}
   \end{align}

Thus, we have demonstrated that, starting from an arbitrary discrete valued real vector and sampling from a known uniform distribution, one can obtain a multinoulli random variable whose parameters become a function of the chosen arbitrary discrete vector. Since this effectively represents the mode prior of the latent space, we demonstrate later that one can backpropagate through this fixed functional node and learn $\boldsymbol{\alpha}$ and thereby the mode priors of the latent space. From an implementation perspective, we approximate $\sigma_h$ with a hard sigmoid to enable backpropogation.

\subsection{Noise engineered mode matching}


In this section, we lay down the necessity and sufficiency needed to induce the modal properties of the latent space on the generated data space. A generator $g$ is a non-linear function of the latent variable $\rvz$ that is tasked to produce the desired data samples $\rvx$.
It effectively serves as a sampler for $P_{X}$, the data generating distribution.

  We desire that $P_{X}$ is a multimodal distribution with $M$ modes and let $\sX_i\subseteq \sX$ represent $i^{\text{th}}$ mode of $P_{X}$. In the following lemma, we state the necessity on $ \rvz $ for  $P_{X}$ to be mumtimodal. 

\textbf{Lemma 2.} If $ \sZ_i\subseteq \sZ$ denote the inverse images of ${\sX}_i$ under $g$, then $ \bigcap_i{\sX}_i = \Phi$ only if $ \bigcap_i{\sZ}_i = \Phi$, where $\Phi$ is an empty set. 

\textit{Proof: } For simplicity we assume $M = 2$, albeit all the analysis holds equally well for larger $M$. Assume $\displaystyle \sX_0 \cap \sX_1 = \Phi$ and  $\displaystyle \sZ_0 \cap \sZ_1 \neq \Phi \implies  \exists \rvz_i \in \sZ_0 \cap \sZ_1$.
Given $\displaystyle \rvz_i \in \sZ_0$, let $g(\rvz_i) = \rvx_{i0} \in \sX_0$ and similarly, given $\displaystyle \rvz_i \in \sZ_1$, $g(\rvz_i ) = {\rvx}_{i1} \in {\sX}_1$. Since $g$ is a continuous function, ${\rvx}_{i0} = {\rvx}_{i1} = {\rvx}_{i} \implies {\rvx}_{i} \in \sX_0 \cap \sX_1$ contradicting the fact that $\sX_0 \cap \sX_1 = \Phi$, hence $\displaystyle \sZ_0 \cap \sZ_1 = \Phi.$ 


From Lemma 2, it is imperative that to obtain a multimodal generated distribution, it is necessary to have a multimodal latent distribution, however it is not sufficient since the generating function $g$ can be non-injective. However, if there exists another mapping $ h:{\sX}\rightarrow {\hat{\sY}}$ which maps the co-domain of $g$, to another random variable $\mathbf{\hat{y}}$ such that $P_{\hat{Y}}$ is also multimodal, then Lemma 2 to be applied again on $h$ and achieve multimodality on $P_{X}$. The following corollary to Lemma 2 is stated to affirm this fact. 

\textbf{Corollary 2.1. }
Let  $ h:\sX\rightarrow \hat{\sY}$ and $\hat{\sY}_i\subseteq \hat{\sY}$ be a subset of $\hat{\sY}$. Then $ \bigcap_i\hat{\sY}_i  = \Phi$ only if $ \bigcap_i{\sX}_i  = \Phi$.  Given $\bigcap_i{\sZ}_i = \Phi$, $\bigcap_i\hat{\sY}_i = \Phi$ is a sufficient condition for $ \bigcap_i{\sX}_i= \Phi$. 

Corollary 2.1 states that if latent distribution is multimodal with $M$ modes and $h$ maps $\rvx$ to any multimodal distribution with $M$ modes, the generated distribution ($P_X$) will also have $M$ modes. 
We now show that if  $P_{Y}$ and $P_{\hat{Y}} $  are enforced to be close, then the modal properties of $P_X$ are matched to $P_Z$, and $P_Y$.

\textbf{Lemma 3.}  Let $\hat{\rvx}$ be a discrete random variable  that is an indicator of modes of $P_X$. That is,  $P_{\hat{\sX}}(\hat{\rvx} =i) = \int_{{\sX}_i}P_X~d\rvx$. Then minimization of KL divergence, $\KL (P_{\hat{Y}} ||P_{Y} )$, is equivalent to minimization of $\KL (P_{\hat{Y}}  || P_{\hat{X}})$. 

\textit{Proof:} 
\begin{align}
\KL (P_{\hat{Y}} || P_{Y}) &= \sum_{\hat{\rvy}=i} {P_{\hat{Y}}}\log \frac{P_{\hat{Y}}}{P_Y} \quad\quad s.t. \quad i\in \{0, 1\} 
\label{kl1}\\
&= \sum_{\hat{\rvy}=i} \left({P_{\hat{Y}}}\log {P_{\hat{Y}}} - {P_{\hat{Y}}}\log {P_Y}\right)
\label{kl2}\\
&= \sum_{\hat{\rvy}=i} \left({P_{\hat{Y}}}\log {P_{\hat{Y}}} - {P_{\hat{Y}}}\log {\int_{\sZ_i}P_{Z}~d\rvz} \right)
\label{kl4}
\end{align}
Since $\int_{ \sZ_i }P_{Z} ~d\rvz = \int_{\sX_i}P_X~d\rvx$, \eqref{kl2} can be written as
\begin{equation}
\KL (P_{\hat{Y}} || P_{Y}) = \sum_{\hat{\rvy}=i} \left({P_{\hat{Y}}}\log {P_{\hat{Y}}} - {P_{\hat{Y}}}\log {\int_{\sX_i}P_X~d\rvx} \right)
\label{kl5}
\end{equation}
Since  $\int_{{\sX}_i}P_X~d\rvx = P_{\hat{X}}(\hat{\rvx} =i )$, by definition, \eqref{kl5} can be written as
\begin{align}
\KL (P_{\hat{Y}} || P_{Y}) &= \sum_{\hat{\rvy}=i} \left({P_{\hat{Y}}}\log {P_{\hat{Y}}} - {P_{\hat{Y}}}\log {P_{\hat{X}}}\right)
\label{kl6.1}\\
&= \KL (P_{\hat{Y}} || P_{\hat{X}})
\label{kl7} 
\end{align}

\textbf{Corollary 3.1. }
Minimizing $\KL (P_{\hat{Y}}|| P_{Y})$ is equivalent to minimizing $\KL (P_{\hat{Y}}|| P_{\hat{X}})$ and thus leads to matching of modal properties of $P_{\hat{X}}$ and $P_Y$.



Lemma 3 and Corollary 3.1 state an important fact that modal properties of the latent space and the data generating space can be matched by choosing $h$ which would minimize  $\KL (P_{\hat{Y}} ||P_{Y} )$. This also implies that the imbalance (if any) in the modes of the latent distribution is reflected in the data generating distribution.  This is especially useful for datasets containing imbalanced classes, which is often the case in real-world applications. Using Lemma 2 and 3, one can make the data generating distribution to be multimodal, however, produced modes might be degenerated in a sense that $\sX_i$'s could be reduce to singletons (intra mode collapse).  To avoid this degenerative case, we propose to decompose  $h$  as a composite of two mappings $h_1:{\sX}\rightarrow \hat{\sZ}$ and $h_2:\hat{\sZ}\rightarrow \hat{\sY}$. Minimizing a norm distance between the samples of $\sZ$ and $\hat{\sZ}$ prevents degenerative modes in $P_X$. This is because $h_1$ enforces a unique reconstruction of every sample of $\rvz$ which in turn ensures that a unique sample of $\rvx$ is generated by a unique sample of $\rvz$. The function $h_1$ can be seen as an activity regularizer that would force every unique noise sample within each mode to map to a unique sample in the inversion and the generated spaces.


\subsection{Realization using GAN}
\label{GAN}
In the above formulation, the generated distribution is not constrained to be close to the distribution of the true data. For that, one has to rely on adversarial training of GANs which would enforce $g(\rvz)$ to be close to the distribution of the true data. Thus training a GAN within the proposed formulation, simultaneously enforces the generated data distribution to be close to the true data distribution and match the modal properties of the latent distribution. 
Conversely, if the latent distribution is chosen in accordance with the modal properties of the true data distribution, generated data space will be clustered conditioned on the modes of the true data i.e., the mode probabilities in generated data are matched with the true data mode probabilities. Hence we propose to realize $g$ as the Generator of a GAN with the usual discriminator network $d$, and $h$ as a neural network operating on the output of the generator. Thus the objective function optimizes over $(g,h_1,h_2,d)$ and given as follows:
\begin{equation}
    \min_{g, h_2\circ h_1} \max_d \mathcal{L}\left(g, h_1, h_2, d\right)
    \label{obj}
\end{equation}
\vspace{-0.5cm}
\begin{multline}
    \hspace{-0.3cm}\mathcal{L}\left(g, h_1, h_2, d\right) =  \E_{\rvx_r} [ \log d(\rvx_r) ] + \E_{\rvz} [ \log\left( 1 - d \circ g(\rvz) \right) \\ + || \rvz - h_1 \circ g(\rvz) ||_p] + \KL (P_{\hat{Y}} || P_{Y})
    \label{obj_l}
\end{multline}
where $\rvx_r$ represents samples from the true data distribution. Figure.~\ref{bd} depicts an overview of the proposed formulation. It can be seen that a chain of continuous functions of random variables   $\boldsymbol{\alpha} \rightarrow\rvy\rightarrow \rvz\rightarrow \rvx \rightarrow \hat{\rvz} \rightarrow  \hat{\rvy}$ is established while imposing a similar modal structures on all of them.  Thus, we call our approach Noise Engineered Mode-matching GAN (NEMGAN).


NEMGAN offers several advantages compared to a vanilla GAN - (a) Once trained, sampling $\rvz$ from a particular mode confines $\rvx$ to a unique mode, leading to conditional generation, (b) when training by imposing a fixed prior on $\rvz$, unseen attributes in the data can be discovered, (c) post training, the $h$ network can be used independently to cluster the data, (d)  Mode-collapse is discouraged in our formulation, since $g$ is explicitly forced to produce a multimodal distribution. 


\subsection{Learning the mode priors}

In the discussions thus far, the priors on the latent space modes parameterized by $\boldsymbol{\alpha}$ vector, was held a known constant. However, often in real-world cases, the data inherently comes with imbalanced modes or classes with unknown imbalance/modal ratios. It is reasonable to assume that one has knowledge about number of modes  $(M)$ in the data (as with the case of K-means and GMM) but not the mode masses. The most natural and commonly adopted way of constructing $P_Y$ with uniform mode priors~\cite{chen2016infogan,mukherjee2018clustergan}, suits well for data whose modes have uniform masses. However, Lemma 2 clearly shows that a $P_Y$ with uniform modal masses will lead to uniformly generated modes and the imbalance in true data cannot be replicated in the generated space $\sX$, leading to a poor generation. Thus, constructing the multimodal latent space in accordance to the data prior is the key to mode matching. Thus we propose a technique to learn the priors on the latent space (optimal $\boldsymbol{\alpha}$ vector) so that it follows the true data priors.
The bottleneck here is that entire pipeline of NEMGAN has no signal to indicate that the assumed prior ($\boldsymbol{\alpha}$) is incorrect. This is because the generator, discriminator and the inverter are agnostic to the idea of a mode since there is an inherent subjectivity in the definition of a mode. 
Though one can argue that there will be `natural' modes in the data distribution, the modes discovered in a completely unsupervised manner may or may not be of practical importance (e.g. image clustering on the basis of background color). Thus, as observed in \cite{locatello2018challenging}, it is impossible to infer the modes and the corresponding priors of the true data distribution in a completely unsupervised way.  However, if there is access to the mode-labels of very few samples (less than one percentage) of the true data (obtained through annotations) then we show, as follows, that the priors of the latent space can be learned to match that of the true data. 
The latent space inverter network $h(\rvx)$ is an estimator of the posterior of the modes given the data, $P(\hat{\rvy}|\rvx)$. Thus marginalizing the output of $h(\rvx)$ over all $\rvx$ amounts to computing  $\E_{\rvx}[h(\rvx)] = P_{\hat{Y}}$.  If the initial assumptions on the latent mode prior is incorrect, then the modes of the generated data learned by the GAN does not match with the true data. This implies that the inferred modes by $h(\rvx_{cr})$, $P(\hat{\rvy}|\rvx_{cr})$ on a few of $\rvx_{cr}$ will be incorrect, where $\rvx_{cr} \in \sX_{cr}$ denoting a small subset sampled from true data distribution with known modes. At this stage, we propose to treat $h(\rvx)$ as an independent network and train it (using a classification loss such as cross entropy) to correctly estimate the modes on $\sX_{cr}$. This further leads to a new estimate of $P_{\hat{Y}}$ say $\hat{P}_{\hat{Y}}$, on any arbitrary subset of data since it is obtained from a different (retrained) $h$ network. Subsequently, we use this mismatch between the aggregated mode posteriors to adjust the value of $\boldsymbol{\alpha}$ by backpropogating the $\KL (P_{\hat{Y}} ||\hat{P}_{\hat{Y}})$ while keeping the other network parameters fixed. At convergence, the mode priors obtained using $\boldsymbol{\alpha}$ are matched with the true data mode priors. Therefore in this case of prior learning, the objective of the NEMGAN  (\eqref{obj}) is updated as
\begin{equation}
    \min_{g, h_2\circ h_1} \max_d \mathcal{L}\left(g, h_1, h_2, d\right) + \min_{h}\mathcal{L}_{cc} + \min_{\boldsymbol{\alpha}}\KL (P_{\hat{Y}} ||\hat{P}_{\hat{Y}})
\end{equation}
where $\mathcal{L}_{cc}$ is the categorical crossentropy loss used to train $h$ on $\sX_{cr}$.



\section{Experiments}
We consider MNIST~\cite{lecun2010mnist}, FMNIST\footnote{https://github.com/zalandoresearch/fashion-mnist}, CelebA~\cite{liu2015deep}, and Stacked MNIST~\cite{srivastava2017veegan} datasets for experiments. We first performed the experiments to evaluate NEMGAN's capability for mode discovery and matching on MNIST, FMNIST, and CelebA with uniform and skewed modes. We compared the observations with other GANs having the mode matching characteristic and tractable latent posterior, including CaTGAN~\cite{springenberg2015unsupervised}, InfoGAN~\cite{chen2016infogan}, GANMM~\cite{yu2018mixture}, and ClusterGAN~\cite{mukherjee2018clustergan}. Subsequently we used Stacked MNIST to evaluate the performance of NEMGAN in presence of large number of clusters and compared it with InfoGAN, BEGAN~\cite{berthelot2017began}, Unrolled GAN~\cite{metz2016unrolled}, and MAD-GAN~\cite{ghosh2017multi}. Finally we conduct a curious experiment of discovering the unknown attributes of the data and share some interesting observations.

The proposed method is generic and can be applied to any GAN to convert it into NEMGAN. We use DCGAN~\cite{radford2015unsupervised} model and modified it to build NEMGAN for MNIST and Stacked MNIST datasets\footnote{Implementation code - https://github.com/NEMGAN/NEMGAN-P}. For more realistic datasets, FMNIST and CelebA, we adopt the improved WGAN~\cite{gulrajani2017improved}. We use Adam optimizer ($\beta_1 = 0.5, \beta_2 = 0.9$) with learning rates of $2\times10^{-4}$ for generator and $10^{-3}$ for both discriminator and inversion network.
For existing methods the codes are obtained from sources made available by the respective authors. In order to quantitatively compare the performance of the methods considered, we use three metrics namely, clustering accuracy (ACC), normalized mutual information (NMI) and adjusted rand index (ARI), on balanced test data, with respect to the latent space reconstruction network~\cite{fahad2014survey}. Higher values of all three metrics represent the modes generated with higher purity, thereby provide a measure of mode matching. 
However, these metrics alone cannot fully capture the quality of the generated images, therefore, we also report Frechet Classification Distance (FCD) for the models. A low FCD implies that the generated data is very close to the true data. Jointly, a low FCD value and high values of the clustering metrics (ACC, NMI and ARI) imply that the generated data is close to the true data and the generated modes are matched with the true data modes. Note, however, that FCD values are highly dependent on the architecture used for the generator and getting low values for FCD, while desirable, was not the primary aim of these experiments. The FCD value is only used to quantify the relative image quality of different models under an identical training procedure.


\subsection{Standard mode discovery}
\begin{figure}[!t]
\centering
\includegraphics[width=4cm]{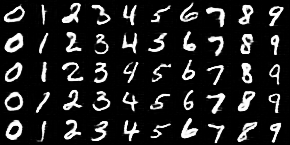}
\includegraphics[width=4cm]{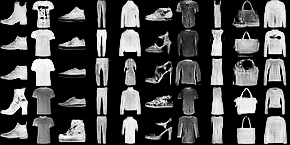}
\caption{Samples generated by NEMGAN$_\text{P}$ for MNIST (left) and FMNIST (right) data. Each column represents one mode. The generated modes are not only pure, they also follow the order imposed during supervision.}
\label{std}
\end{figure}

We begin with the experiments on MNIST and FMNIST datsets to discover the standard 10 modes in both sets and perform mode matching. We train three versions of the proposed GAN, namely NEMGAN$_\text{V}$, NEMGAN$_\text{S}$, and NEMGAN$_\text{P}$. NEMGAN$_\text{P}$ represents the model in which priors are learned using $<1\%$ of true data along with mode matching, NEMGAN$_\text{S}$ represents the model in which $<1\%$ of true data is used to supervise $h$ network with no prior learning and NEMGAN$_\text{V}$ represents a vanilla model in which neither prior learning nor supervision of $h$ is considered, NEMGAN$_\text{V}$ is trained to match the modes assuming uniform priors. Quantitative evaluation of all these models is summarized in Table~\ref{tab:10class}. For MNIST, which contains well-separated and uniformly distributed natural modes, NEMGAN$_\text{V}$ and NEMGAN$_\text{P}$ gives the best performance, whereas for FMNIST, where the mode priors are not uniform, NEMGAN$_\text{P}$ outperforms all the other methods including the variants of NEMGAN with a considerable margin. NEMGAN$_\text{S}$, which receives same supervision as NEMGAN$_\text{P}$ but does not learn the mode priors, shows comparatively lower performance. This affirms the necessity of the prior learning for an accurate mode matching. Few samples generated using NEMGAN$_\text{P}$ are shown in Figure~\ref{std}. Note that the generated modes are not only pure, they also follow the order imposed during supervision, which is a side-benefit of the proposed method.

\begin{table}[]
\begin{tabular}{llrrrr}
\hline
\textbf{Dataset} & \textbf{Model} & \textbf{ACC} & \textbf{NMI}       & \textbf{ARI}  & \textbf{FCD}     \\ \hline
& CaTGAN   &0.89&0.90&0.84& 7.34\\
& InfoGAN        & 0.89         & 0.86               & 0.82     &   14.74   \\ 
& GANMM          & 0.64      & 0.61            & 0.49     &     10.83   \\ 
MNIST & ClusterGAN     & 0.95         & 0.89               & 0.89     &   1.84       \\ 
& NEMGAN$_\text{V}$         & \bf 0.96      & \bf 0.91 & \bf 0.92  &  1.82 \\ 
& NEMGAN$_\text{S}$ & 0.96     & 0.90 & 0.91  &  2.25\\ 
& NEMGAN$_\text{P}$  & \bf 0.96      & \bf 0.91 & \bf 0.92 & \bf 1.69\\ \hline
& CaTGAN   & 0.55 & 0.60 & 0.44 & 6.95 \\
& InfoGAN        & 0.61         & 0.59                & 0.44& 12.44          \\ 
& GANMM        & 0.34      & 0.27            & 0.20        & 19.80     \\
FMNIST & ClusterGAN     & 0.63         & 0.64               & 0.50    & 0.56              \\ 
& NEMGAN$_\text{V}$         & 0.65      & 0.61 & 0.53 & 0.55 \\ 
& NEMGAN$_\text{S}$         &   0.81   & 0.70 & 0.62 &  0.58\\ 
& NEMGAN$_\text{P}$        & \bf 0.84      & \bf 0.72 & \bf 0.64 & \bf 0.54\\ \hline
\end{tabular}
\caption{Quantitative evaluation of NEMGAN on MNIST and FMNIST datsets for standard mode detection and matching.}
\label{tab:10class}
\end{table}

\subsection{Mode matching with imbalanced modes}

\begin{figure}[!t]
\centering
\begin{minipage}[]{0.48\linewidth}
  \centerline{\includegraphics[width=4cm]{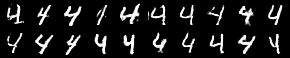}}
\end{minipage}
\begin{minipage}[]{.48\linewidth}
  \centerline{\includegraphics[width=4cm]{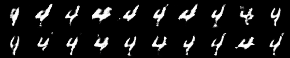}}
\end{minipage}

\begin{minipage}[]{0.48\linewidth}
  \centerline{\includegraphics[width=4cm]{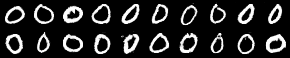}}
  \centerline{\footnotesize{(a) 25:75} }
\end{minipage}
\begin{minipage}[]{.48\linewidth}
  \centerline{\includegraphics[width=4cm]{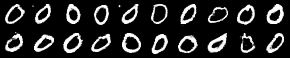}}
  \centerline{\footnotesize(b) 02:98}
\end{minipage}
\caption{Sample images generated in the experiments with different class ratios. Surprisingly the network is able to do conditional generation even with an imbalance of 02:98.  }
\label{gen_im0_4}
\end{figure}

\begin{figure}[!t]
\centering
\begin{minipage}[]{0.48\linewidth}
  \centerline{\includegraphics[width=4cm]{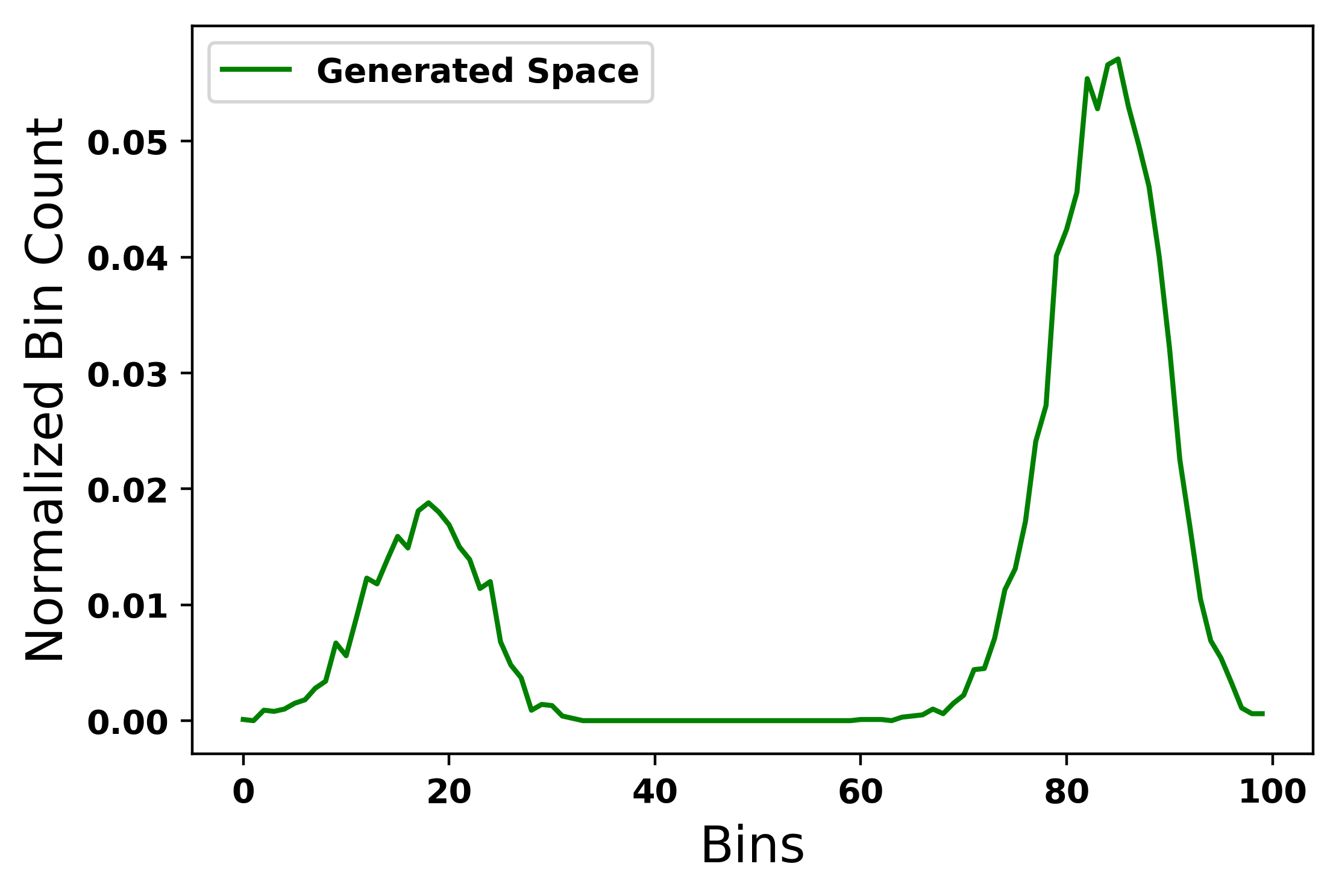}}
\end{minipage}
\begin{minipage}[]{.48\linewidth}
  \centerline{\includegraphics[width=4cm]{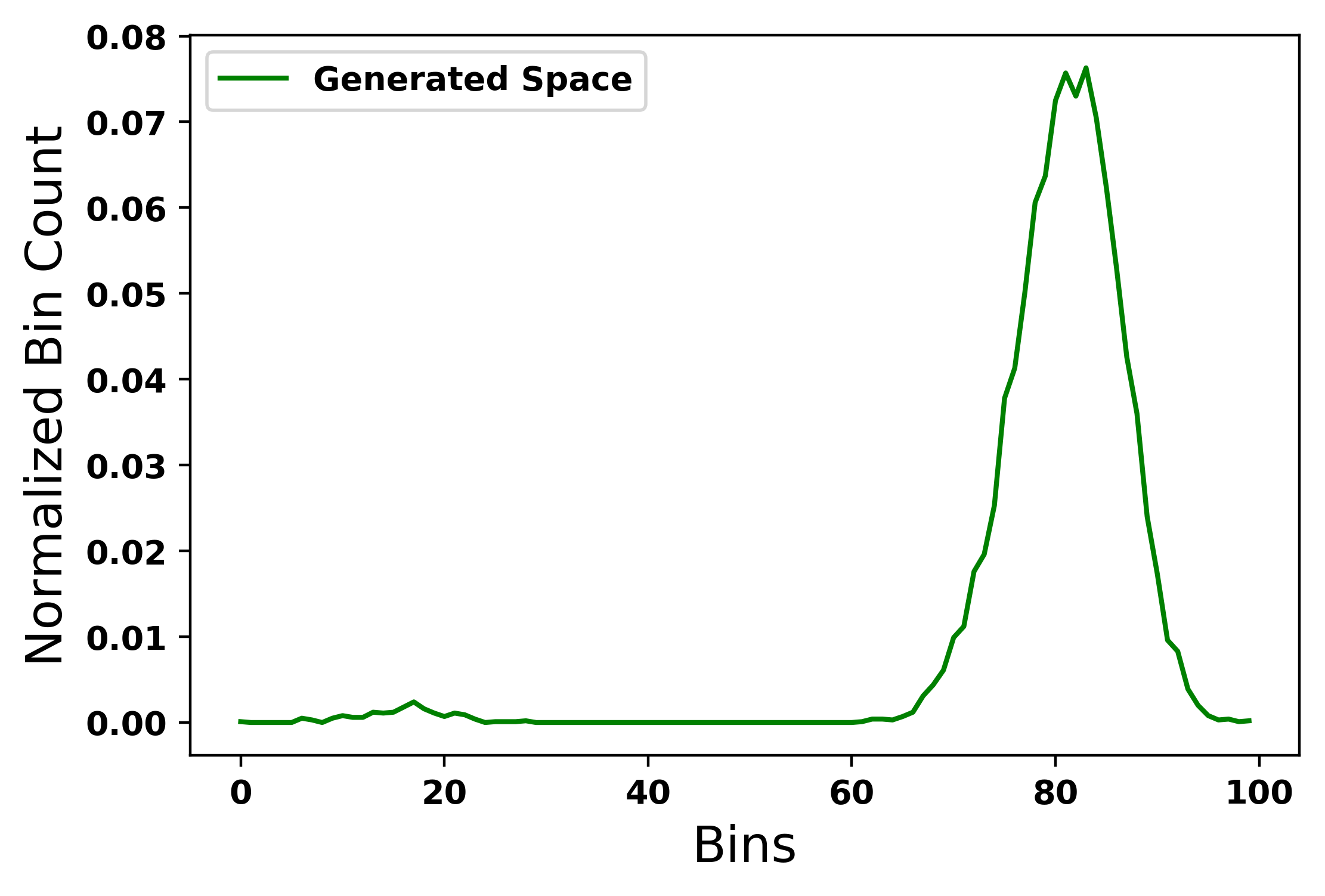}}
\end{minipage}

\begin{minipage}[]{0.48\linewidth}
  \centerline{\includegraphics[width=4cm]{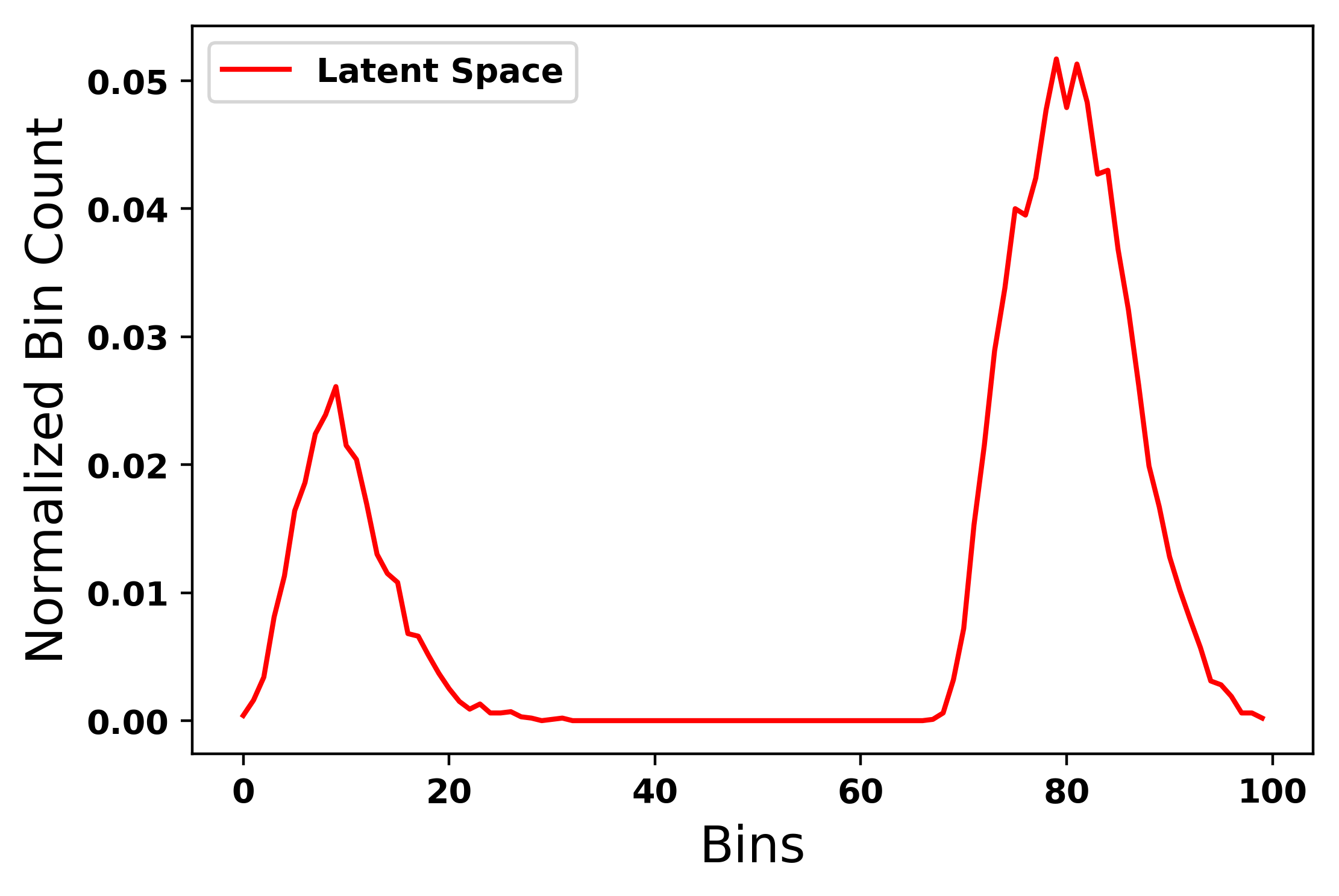}}
  \centerline{\footnotesize{(a) 25:75} }
\end{minipage}
\begin{minipage}[]{.48\linewidth}
  \centerline{\includegraphics[width=4cm]{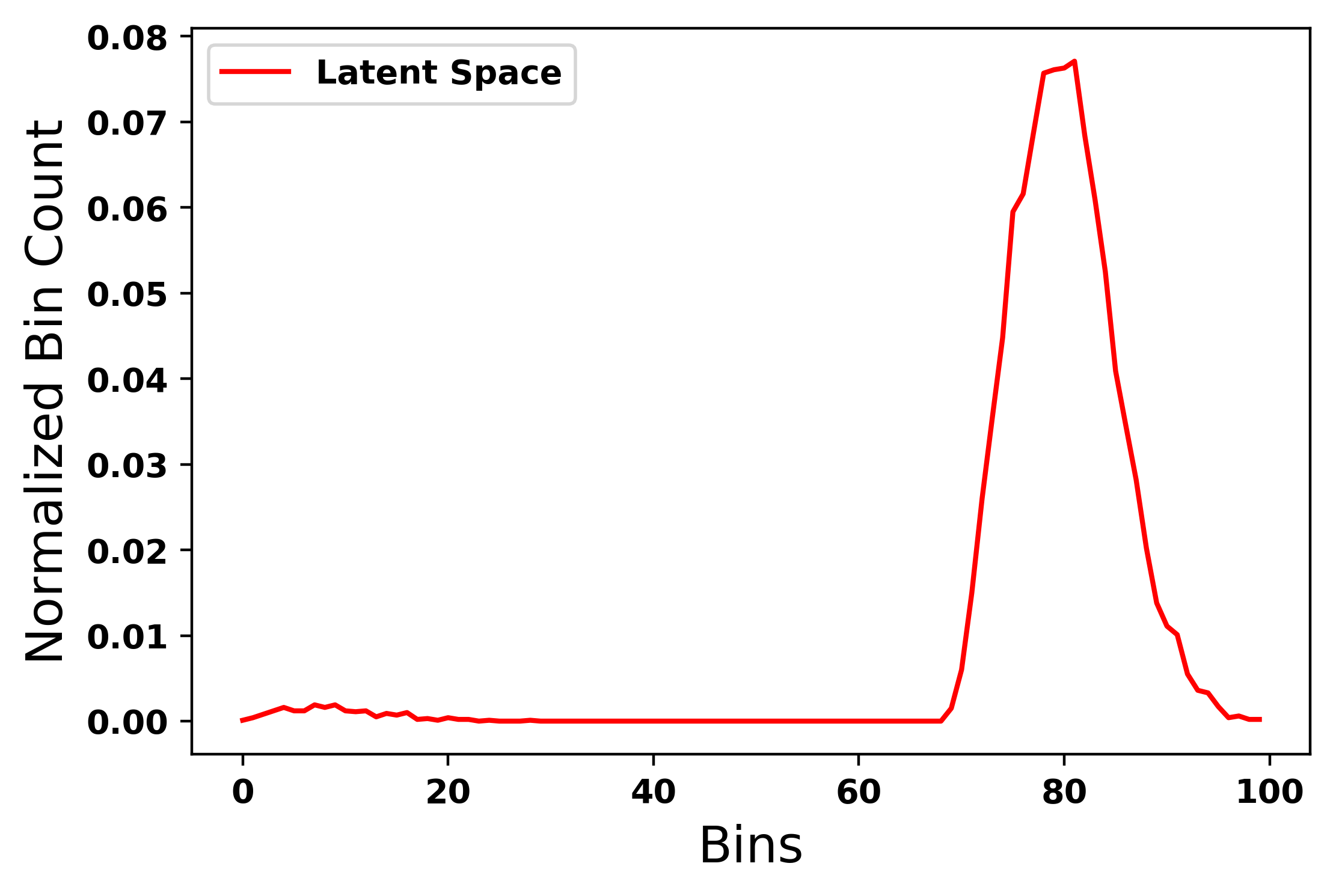}}
  \centerline{\footnotesize(b) 02:98}
\end{minipage}
\caption{Distribution of sample distances from the corresponding class means for different ratios. The plots indicate mode matching between the latent and generated data space.}
\label{dist0_4}
\end{figure}

Next we consider the mode matching experiments on imbalanced data, which are the most critical and probably the most important for real-world applications. We train NEMGAN$_\text{P}$ to evaluate its ability to learn the mode priors or imbalance and match the generated data modes with the true data modes. We first take two distinct MNIST classes (digits 0 and 4), identified based on the t-SNE plots and introduce class imbalances by removing samples from one class while retaining all the samples of the other class. Figure \ref{gen_im0_4} depicts the images generated by NEMGAN$_\text{P}$ corresponding to the two modes. It can be seen that even with an imbalance ratio of 02:98, the network is able to conditionally generate semantically separated images, with great variety. Further to illustrate the mode matching, we plot the class-wise histograms of the distances between the individual samples and their corresponding means in Figure~\ref{dist0_4}, for both the noise and the generated data space. It can be seen that there is clear evidence of mode matching between latent and generated spaces.

\begin{table}[]
\begin{tabular}{lp{1.6cm}rrrr}
\hline
\textbf{Dataset} & \textbf{Model} & \textbf{ACC} & \textbf{NMI}       & \textbf{ARI}  & \textbf{FCD}     \\ \hline
& CaTGAN    & 0.95 & 0.75 & 0.81 & 7.21   \\
& InfoGAN  & 0.73   & 0.16  & 0.21     & 5.39 \\ 
MNIST-2 & GANMM & 0.66      & 0.08  & 0.11     &   19.07     \\ 
(50:50) & ClusterGAN     & \bf 0.99 & 0.91   & 0.95     &  1.51        \\
& NEMGAN$_\text{P}$  & \bf 0.99 & \bf 0.92 & \bf 0.96 & \bf1.47\\ \hline
& CaTGAN    & 0.66 & 0.07 & 0.11 & 10.14\\
& InfoGAN  & 0.75   & 0.20  & 0.25     &     15.11     \\ 
MNIST-2 & GANMM & 0.64      & 0.06  & 0.08     &    19.76  \\ 
(70:30) & ClusterGAN     & 0.94 & 0.72   & 0.79     &    1.56  \\
& NEMGAN$_\text{P}$  & \bf 0.98      & \bf 0.89 & \bf 0.93 & \bf 1.33\\ \hline
& CaTGAN   & 0.59 & 0.05 & 0.03 & 11.45\\
& InfoGAN    & 0.61  & 0.04   & 0.05     &   10.84    \\ 
MNIST-2 & GANMM          & 0.64      & 0.09            & 0.07     &     20.32   \\ 
(90:10) & ClusterGAN     & 0.82  & 0.43 & 0.41     &   2.04    \\
& NEMGAN$_\text{P}$  & \bf 0.98      & \bf 0.86 & \bf 0.91 & \bf 1.66\\ \hline
& CaTGAN   & 0.71 & 0.59 &  0.52 & 12.07\\
& InfoGAN  & 0.71 & 0.55 &  0.54 & 15.31         \\ 
MNIST-5 & GANMM          & 0.51 &0.21&0.19     &    20.64    \\ 
& ClusterGAN     & 0.83         & 0.81               & 0.73     &  1.74        \\
& NEMGAN$_\text{P}$  & \bf 0.96 & \bf 0.89 & \bf 0.89 & \bf 1.13 \\ \hline
& CaTGAN   & 0.77 & 0.66 & 0.61 & 5.41\\
& InfoGAN  & 0.77 & 0.58 & 0.55    &    17.20      \\ 
FMNIST-5 & GANMM  & 0.62 & 0.30 & 0.30    &     25.46   \\ 
& ClusterGAN     & 0.81         & 0.66  & 0.62     &    2.39      \\
& NEMGAN$_\text{P}$  & \bf 0.92 & \bf 0.81 & \bf 0.81 & \bf 0.69 \\ \hline
& CaTGAN   & 0.58 & 0.05 & 0.04 & 67.1 \\
& InfoGAN  & 0.57 & 0.04 & 0.03   & 110.9 \\ 
CelebA & GANMM      & 0.55 & 0.02 & 0.01 & 250.2   \\ 
& ClusterGAN       & 0.76 & 0.19 & 0.22   & 73.4        \\
& NEMGAN$_\text{P}$ & \bf 0.81 & \bf 0.30 & \bf 0.38 & \bf  62.9 \\ \hline
\end{tabular}
\caption{Quantitative evaluation of NEMGAN on imbalanced data for mode matching. We train the GANs with an equal amount of minimal-supervision as NEMGAN$_\text{P}$, however, in absence of prior learning considerably lower performances are observed.}
\label{tab:imbalance}
\end{table}

Next we consider an imbalanced combination of digits 3 and 5 (MNIST-2), which have a significant overlap in the t-SNE plots, and test NEMGAN$_\text{P}$ on these. We vary mode ratios of class 3 and 5 from 50:50 to 90:10. We compare NEMGAN$_\text{P}$ with the existing approaches and for a comprehensive evaluation, we provide an equal amount of supervision as the proposed method to the existing approaches as well. The observed evaluation metric values are listed in Table~\ref{tab:imbalance}. As the mode skew increases, the performance of all the existing methods degrades. In contrast NEMGAN shows a consistent performance and outperforms the other approaches with a considerable margin for the ratio of 90:10. 

\begin{figure}[!t]
\centering
\includegraphics[width=2.5cm]{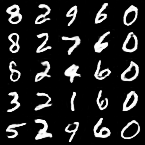}
\hspace{0.5cm}
\includegraphics[width=2.5cm]{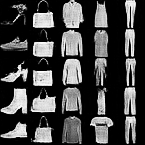}
\caption{Samples generated by NEMGAN$_\text{P}$ for MNIST-5 (left) and FMNIST-5 (right) data. Each column represents one mode. The generated modes clearly capture the grouping of the classes.}
\label{sample5}
\end{figure}

We further extend these experiments to all the 10 digits, however, instead of introducing an artificial imbalance, we grouped together similar digits \{\{3,5,8\}, \{2\}, \{1,4,7,9\}, \{6\}, \{0\}\} to form a 5-class MNIST dataset. A natural imbalance is observed due to the grouping of the digits. Similarly, we also grouped FMNIST classes to create the FMNIST-5 dataset as {\{Sandal, Sneaker, Ankle Boot\}, \{Bag\}, \{Tshirt/Top, Dress\}, \{Pullover, Coat, Shirt\}, \{Trouser\}}. Performance of all the considered methods is summarized in Table~\ref{tab:imbalance}. In these experiments also, we train the existing GANs with minimal-supervision same as NEMGAN$_\text{P}$, however, in absence of prior learning comparatively lower performances are observed. Figure~\ref{sample5} shows few samples generated by NEMGAN$_\text{P}$ for MNIST-5 and FMNIST-5 datasets. Note that in absence of any supervision the networks have no clue about the grouping. This is also observed by~\cite{mukherjee2018clustergan}, where any experiment on MNIST data with number of modes different from 10 showed performance degradation. Thus, the minimal-supervision provides a way to supply the user defined knowledge of modes to NEMGAN, which is an additional benefit of the proposed method.

For a more realistic experiment we consider CelebA dataset and train NEMGAN$_\text{P}$ to distinguish celebrities with black hair from the rest. There are only 23.89\% celebrity pictures with black hair in the complete datset, which indicates the imbalance in the desired modes. These experiments provide empirical evidence to the mode matching in real scenarios and also affirm the requirement for prior learning and minimal-supervision provided to NEMGAN$_\text{P}$ using $<1\%$ of true data samples. 

\subsection{Mode counting}
Although, the aforementioned experiments demonstrate NEMGAN's ability to discover as well match the modes, we additionally consider Stacked MNIST dataset with 1000 modes to evaluate the performance in presence of large number of clusters. As mentioned in \cite{ghosh2017multi} we create stacked MNIST dataset by stacking three random MNIST digits along color channels of an RGB image to obtain color images. This dataset can have 1000 modes corresponding to every possible combination of digit triplets. We used experimental settings identical to~\cite{ghosh2017multi} which allows us to obtain the results of the existing methods from~\cite{ghosh2017multi}. Table~\ref{tab:stack} lists the results of our method and the other methods on discovering the modes. For our method, latent space is a discrete uniform distribution with 1000 modes. It can be seen that, NEMGAN is able to discover all 1000 modes, which is remarkable.

\begin{table}[]
\centering
\begin{tabular}{lcc}
\hline
\textbf{Model} & \textbf{Modes(Max 1000)} & $\boldsymbol{\KL}$   \\ \hline
BEGAN   & 819 & 1.89\\
InfoGAN        & 840         & 2.75 \\ 
Unrolled GAN          & 842      & 1.29    \\ 
MAD-GAN     & 890        & 0.91       \\ 
NEMGAN  & \bf 1000   &  \bf 0.21\\ \hline
\end{tabular}
\caption{Modes captured by different models for the stacked MNIST dataset.}
\label{tab:stack}
\end{table}

\subsection{Attribute discovery}  
\begin{figure}[!t]
\centering
\includegraphics[height=3.7cm]{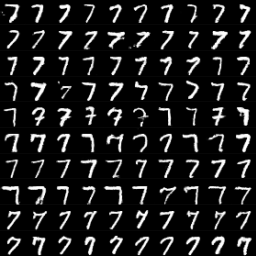}
\includegraphics[height=3.7cm]{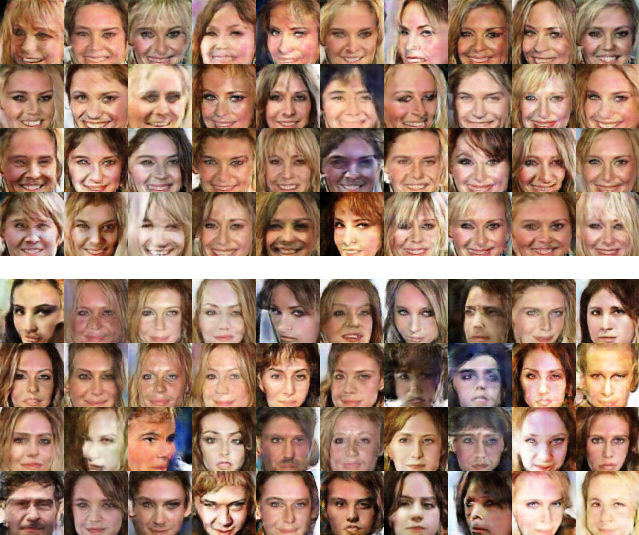}
\caption{Attributes discovered by NEMGAN for 7 from MNIST (left) and CelebA (right) data. Each row represents one mode.}
\label{attribute}
\end{figure}

In a few real-life scenarios, the data comes with no information at all. In such cases, the assumption of knowledge of number of modes and availability of few annotated samples for minimal-supervision become invalid. Thus, the scenario becomes completely unsupervised where the modes and corresponding priors are need to be fixed and the model is expected to discover the plausible attributes of the data.
To evaluate NEMGAN's ability to perform such a task, we use digit 7 of MNIST dataset to divide the images into ten clusters with fixed uniform priors representing different attributes. Similarly we use CelebA dataset and use NEMGAN to discover two modes having a ratio of 30:70. Figure~\ref{attribute} shows the generated samples. For digit 7, each mode, presented in rows, captures a different writing style and stroke. Similarly, in celebA experiment, the network discovers visibility of teeth as an attribute to the faces. Note that these attributes are not present in the available standard labels of the datasets, but are discovered by NEMGAN.

\section{Conclusion and future work}
We construct a framework for mode matching and discovery which can be applied to any GAN formulation. We also propose to  parameterize the latent space and learn the latent priors, using minimal-supervision ($<1\%$), which enables matching of the statistical properties of latent and data distributions. Our method offers several advantages over the existing methods, including robust generation under the setting of skewed data distributions, better mode separation, and benefiting from the user knowledge of true data modes. We believe that this method has significant applications in the field of medical image analysis, security and surveillance, where datasets, in general, have high class imbalance. In future, we will explore the possibility of applying our method to non-image datasets.

{\small
\bibliographystyle{ieee}
\bibliography{nemgan}
}

\end{document}